\documentclass[10pt,twocolumn,letterpaper]{article}

\usepackage[pagenumbers]{cvpr} 

\usepackage{graphicx}
\usepackage{float}  %
\usepackage{graphicx} 
\definecolor{cvprblue}{rgb}{0.21,0.49,0.74}
\usepackage[pagebackref,breaklinks,colorlinks,allcolors=cvprblue]{hyperref}
\def\paperID{27} 
\def\confName{CVPR}
\def\confYear{2025}

\title{Detect, Classify, Act: Categorizing Industrial Anomalies with Multi-Modal Large Language Models}

\author{Sassan Mokhtar\\
University of Bonn\\
Bonn, Germany\\
{\tt\small mokhtar@iai.uni-bonn.com}
\and
Arian Mousakhan\\
University of Freiburg\\
Freiburg, Germany\\
{\tt\small mousakha@cs.uni-freiburg.de}
\and
Silvio Galesso\\
University of Freiburg\\
Freiburg, Germany\\
{\tt\small galessos@cs.uni-freiburg.de}
\and
Jawad Tayyub\\
Endress + Hauser\\
Maulburg, Germany\\
{\tt\small jawad.tayyub@endress.com}
\and
Thomas Brox\\
University of Freiburg\\
Freiburg, Germany\\
{\tt\small brox@cs.uni-freiburg.de}
}

\begin{document}
\maketitle
\begin{abstract}
Recent advances in visual industrial anomaly detection have demonstrated exceptional performance in identifying and segmenting anomalous regions while maintaining fast inference speeds. However, anomaly classification—distinguishing different types of anomalies—remains largely unexplored despite its critical importance in real-world inspection tasks. To address this gap, we propose VELM, a novel LLM-based pipeline for anomaly classification. Given the critical importance of inference speed, we first apply an unsupervised anomaly detection method as a vision expert to assess the normality of an observation. If an anomaly is detected, the LLM then classifies its type. A key challenge in developing and evaluating anomaly classification models is the lack of precise annotations of anomaly classes in existing datasets. To address this limitation, we introduce MVTec-AC and VisA-AC, refined versions of the widely used MVTec-AD and VisA datasets, which include accurate anomaly class labels for rigorous evaluation. Our approach achieves a state-of-the-art anomaly classification accuracy of $80.4\%$ on MVTec-AD, exceeding the prior baselines by $5\%$, and $84\%$ on MVTec-AC, demonstrating the effectiveness of VELM in understanding and categorizing anomalies. We hope our methodology and benchmark inspire further research in anomaly classification, helping bridge the gap between detection and comprehensive anomaly characterization.
\noindent\textbf{Code:} \href{https://github.com/Sassanmtr/VELM}{\textcolor{cvprblue}{github.com/Sassanmtr/VELM}}

\end{abstract}

\section{Introduction}
\label{sec:intro}

\begin{figure}[t]
    \centering
    \includegraphics[width=1\linewidth]{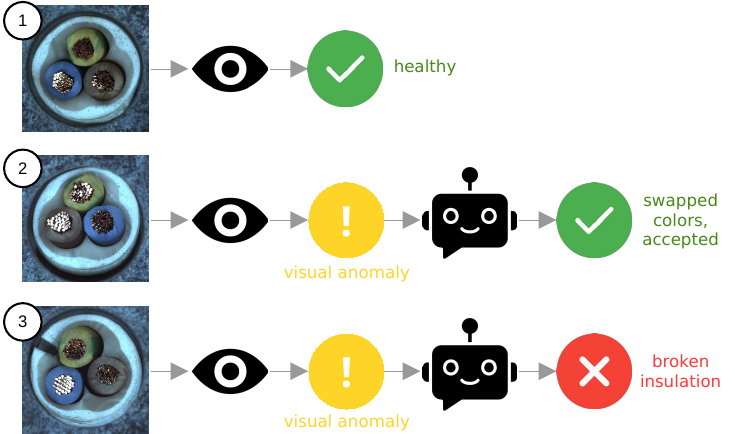}
    \caption{Example of the proposed use case for our anomaly detection and classification pipeline. In most cases (1), the test samples will be directly deemed healthy by a visual detector (e.g. PatchCore, DDAD, etc.). If not, a semantic enabled multi-modal model will decide whether the defect is admissible (2) or not (3), based on user-defined instructions.}
    \label{fig:teaser}
\end{figure}

Anomaly detection is a critical component of various computer vision applications, including industrial inspection \cite{mousakhan2023anomaly, roth2022towards}, medical diagnostics \cite{wolleb2022diffusion}, and autonomous driving \cite{nayal2023ICCV, galesso2024diffusion}. Early and accurate detection of anomalies helps prevent costly failures and improve safety. However, simply detecting an anomaly—i.e., determining whether something is abnormal—often falls short of real-world needs. Effective decision-making in practical settings depends on identifying \emph{what} an anomaly is and \emph{how} it should be managed. This gap is especially evident in industrial inspection, where anomalies of different types can demand vastly different responses or, in some cases, no response at all. 

\begin{figure}[t]
    \centering
    \includegraphics[width=1\linewidth]{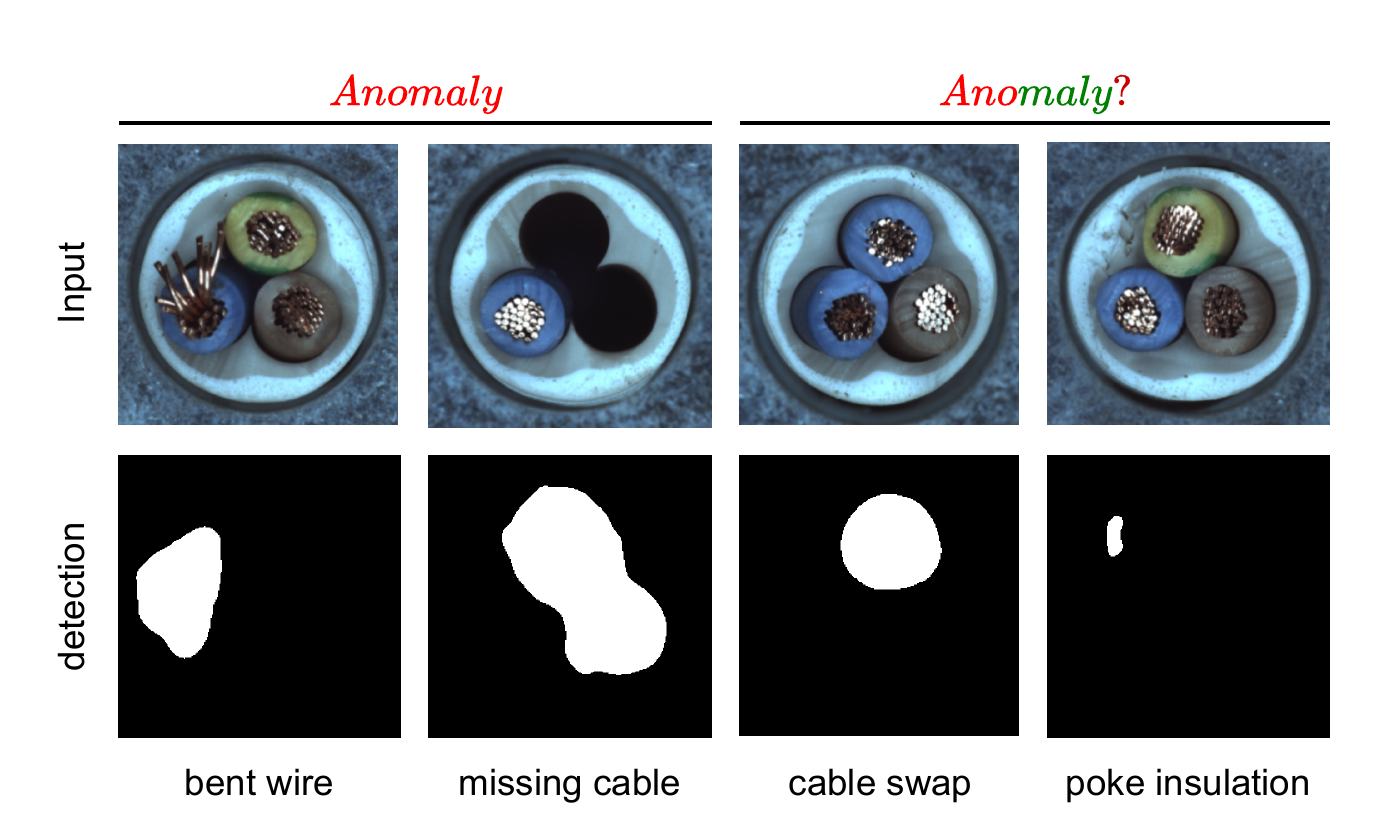}
    \caption{The figure illustrates four anomaly cases detected by a visual anomaly detector \cite{mousakhan2023anomaly}. On the left, clear defects include bent wires and missing cables. On the right, anomalies may not indicate actual faults, such as a color change due to design updates or a minor indentation. Our LLM-based model helps distinguish critical issues from benign variations, ensuring informed decision-making after detection.}
    \label{fig:reallyanomaly}
\end{figure}


Consider the examples in Figure~\ref{fig:reallyanomaly}, where anomalies manifest as: (1)~a bent wire that can be trimmed to salvage production; (2)~missing cables that require more extensive intervention; (3)~a color change that may simply indicate a design update and not a true defect; and (4)~a minor poke that is flagged simply because it was never encountered in training data, despite having negligible impact on functionality. These cases demonstrate that accurate anomaly classification is essential for informed decision-making. We leverage the semantic understanding of Large Language Models (LLMs) to categorize anomalies, enabling context-aware responses in industrial pipelines.

Despite near-perfect accuracy and localization on the common benchmarks, state-of-the-art anomaly detectors remain inadequate for  \emph{anomaly classification}. Two main factors contribute to this limitation. First, these systems rely heavily on visual deviations from a narrowly defined “normal” distribution, causing even benign distribution shifts (e.g., normal design updates) to trigger false positives and necessitate retraining. Second,  their definition of normality is derived solely from the training images, with no integration of broader semantics. However, normality/abnormality is ultimately a user-defined quality, and should be determined through both visual observation and natural language instructions.

Recent developments in large language models (LLMs) and vision-language models (VLMs) offer a promising route to address these challenges, particularly in few-shot and zero-shot anomaly detection scenarios. Due to their extensive pretraining on diverse textual and visual data, LLMs and VLMs can incorporate semantic knowledge that goes beyond the visual norms represented in training sets. Indeed, recent work~\cite{jeong2023winclip} has shown these models can detect anomalies with high accuracy, and further studies~\cite{jiang2024mmad, chen2025can} suggest that with minimal human supervision, they can provide additional context about observed anomalies.

\begin{figure*}[ht]
    \centering
    \includegraphics[width=1\linewidth]{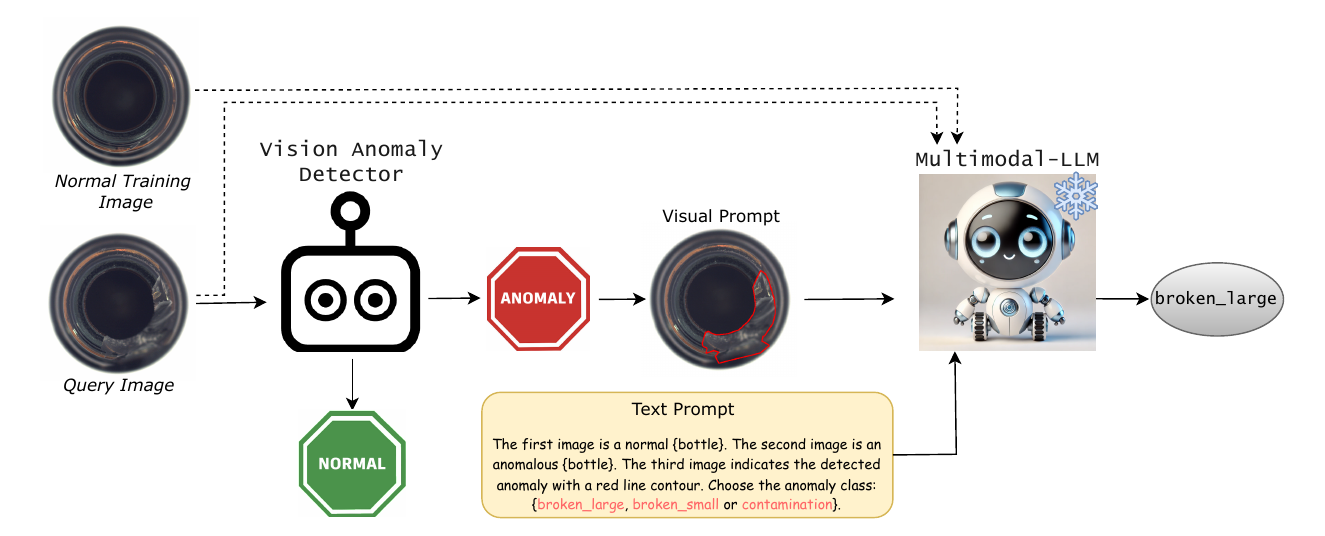}
    \caption{Overview of VELM. Given a query image, VELM first processes it using a Vision Expert, which performs both anomaly detection and localization. If the image is classified as normal, the process terminates. Otherwise, based on the localization from the Vision Expert, a visual prompt is generated by overlaying a red contour on the detected anomaly. The Multimodal-LLM then receives the normal image, query image, visual prompt, and a textual prompt to classify the anomaly into predefined categories}
    \label{fig:framework}
\end{figure*}

Motivated by these insights, we introduce VELM, the first pipeline specifically designed for anomaly classification. As illustrated in Figure~\ref{fig:teaser}, VELM integrates a vision-based anomaly detector with a multi-modal LLM.  The vision expert—any state-of-the-art detector with pixel-level anomaly localization—identifies abnormal samples and highlights the anomalous regions. This ensures a quick normal component detection. In case of an out-of-distribution observation, the anomaly map and expert-designed text prompts are input into the multimodal LLM to classify the detected anomalies. This strategy leverages the speed and accuracy of specialized vision methods while leveraging the semantic richness of LLMs for flexible, context-aware classification.

A further challenge in developing such a system is the lack of properly designed evaluation benchmarks. Existing datasets like MVTec-AD and VisA focus primarily on anomaly detection and localization; although they include anomaly categories, mislabeled samples, and overlapping classes make them unsuitable for comprehensive \emph{classification} evaluation. To overcome these limitations, we refine MVTec-AD and VisA into MVTec-AC and VisA-AC, respectively. These enhanced versions provide precise anomaly class annotations, enabling rigorous and systematic assessment of multi-class anomaly classification methods.

By bridging the gap between anomaly detection and semantic defect characterization, our work suggests anomaly classification as a distinct and essential research direction. We hope that our approach and proposed benchmarks will encourage further exploration in the field, advancing real-world usability and impact of anomaly classification solutions.


\section{Related Work}
\label{sec:related}

\subsection{Visual Anomaly Detection}
Anomaly detection methods for visual industrial inspection can be divided into two classes.
Representation-based methods \cite{cohen2020sub, defard2021padim, deng2022anomaly, batzner2024efficientad} leverage pretrained models such as ResNet \cite{he2016deep} to extract the nominal features. During inference, any observation out of the nominal feature distribution is deemed anomalous. 
With the advances in generative modeling, reconstruction-based approaches gained popularity. 
Methods such as \cite{mousakhan2023anomaly, Zhang_2023_ICCV}, use the power of diffusion models \cite{ho2020denoising, song2020score} to directly learn the representation of the given domain. At inference time, the reconstruction loss is considered as an anomaly score. Both representation-based and reconstruction-based methods have shown promising results on benchmarks such as MVTec-AD~\cite{bergmann2019mvtec} and VisA~\cite{zou2022spot}, excelling in speed and precision and making them suitable for real-time applications. However, 
a significant limitation is their sensitivity to data distribution shifts: even minor changes in the input data often necessitate retraining, which is data-hungry and computationally expensive. Furthermore, these methods primarily focus on visual feature comparisons, lacking inherent semantic understanding of the detected anomalies.


\subsection{Semantic Characterization of Anomalies}
The advent of multi-modal large language models~\cite{achiam2023gpt, bai2023qwen} and vision-language models~\cite{radford2021learning, bousselham2024grounding} 
has impacted anomaly detection.
The power of large language models has been leveraged to perform few-shot or zero-shot anomaly detection on existing datasets~\cite{chen2023clip, jeong2023winclip, zhou2023anomalyclip, li2023myriad, gu2024anomalygpt}. The semantic understanding encoded within language models allows identification of anomalies. However, while these methods incorporate language models, they predominantly focus on detecting the presence of anomalies rather than characterizing their attributes or semantic nature. Among these works, AnomalyGPT~\cite{gu2024anomalygpt} proposes an LLM with anomaly description capabilities, but these are untested and require synthetic data for prompt learning. WinCLIP~\cite{jeong2023winclip} uses textual descriptions of defects as an input modality, but does not enable their semantic understanding. Recent efforts~\cite{jiang2024mmad, chen2025can} have explored the use of multimodal LLMs for the retrieval of attributes of defects through question-answering, however they either suffer from the absence of a specialized visual anomaly detector, or do not provide an automated pipeline without the need for test-time human input. Notably MCAD~\cite{li2024mcad} employs relational knowledge distillation for anomaly classification, but its performance and convenience are limited by the lack of the semantic power and controllability of language models. 

In contrast with the methods mentioned above, our approach provides a useful semantic characterization of the detected anomalies, which is accurate, human-controllable, flexible, and doesn't require additional data or training.

\section{Method}
\label{sec:method}

We propose a multi-stage framework, \textbf{VELM} (\textbf{V}ision \textbf{E}xpert + \textbf{L}anguage \textbf{M}odel)), specifically designed for anomaly classification. VELM strategically integrates a vision-based anomaly detector with a multimodal large language model (LLM) to efficiently detect, localize, and categorize anomalies (see Figure~\ref{fig:framework}). Unlike existing approaches that directly pass query images to an LLM or a vision-language model (VLM), our method employs a dedicated vision-based filtering step to enhance efficiency and accuracy. This strategy minimizes false positives, improves localization precision, and ensures computational efficiency before engaging the LLM for anomaly classification.

\subsection{Vision Expert}

The first stage of VELM employs a vision-based anomaly detector, referred to as the \emph{Vision Expert}, responsible for pixel-wise anomaly localization and filtering normal samples before passing them to the LLM classifier. This module plays two key roles:

\begin{itemize}
\item \textbf{Detection and Localization:} The Vision Expert identifies and localizes anomalous regions while filtering out normal images. This filtering significantly reduces false positives—an issue commonly encountered when directly using LLMs for image-based classification. Additionally, normal samples, whose normality is well-defined within the visual training dataset, are quickly processed due to the fast inference of the Vision Expert. By preventing unnecessary LLM processing of normal images, this step optimizes both computational cost and classification accuracy.

\item \textbf{Visual Prompting:} Inspired by \cite{denner2024visual}, we improve the accuracy of anomaly classification by outlining detected anomalies with red-line contours. The annotated image, along with the original and a normal reference image, is then provided as input to the LLM classifier. This structured input format ensures that the LLM receives context-rich visual information.
\end{itemize}

For our experiments, we use DDAD~\cite{mousakhan2023anomaly} as the Vision Expert due to its high detection accuracy and efficiency. DDAD achieves an inference time of 35 ms—significantly faster than LLMs (e.g., GPT-4 requires 96 ms per token)—making it well-suited for real-time industrial applications.

\subsection{Multimodal LLM-based Anomaly Classifier}

For images flagged as anomalous, VELM employs a multimodal LLM-based classifier that integrates visual inputs with structured text prompts for refined anomaly classification. Unlike traditional classifiers trained on fixed distributions, our framework allows dynamic, user-defined classification categories, making it highly adaptable.

\begin{figure}[t]
\centering
\includegraphics[width=1\linewidth]{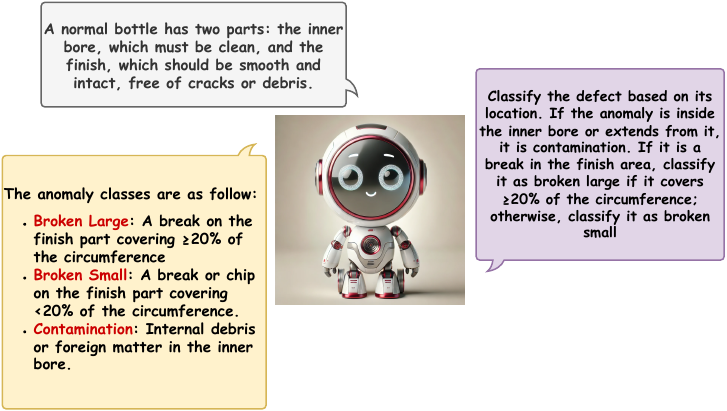}
\caption{Example of a structured text prompt used for anomaly classification with multimodal LLMs. The prompt includes a normal object description, anomaly class definitions, and a classification strategy to guide the model's decision-making}
\label{fig:texdes}
\end{figure}

The classification process is guided by structured prompts (see Figure~\ref{fig:texdes}), consisting of:
\begin{enumerate}
\item \textbf{Normal Object Description:} Detailed descriptions of typical object features (e.g., shape, color, form) to establish a normality baseline.
\item \textbf{Anomaly Class Descriptions:} Clear definitions of all anomaly classes within the dataset, addressing potential ambiguities (e.g., "crack" may have different meanings across object categories).
\item \textbf{Classification Strategy:} Explicit instructions directing the LLM to focus on critical features, ensuring consistent and reliable classification.
\end{enumerate}

By restricting the LLM’s analysis to pre-filtered anomalous samples, our framework ensures that classification is informed by both semantic knowledge and precise localization, enhancing interpretability. Unlike traditional vision-based methods that detect anomalies based solely on data deviations, our approach enables nuanced decisions, such as assessing anomaly severity (e.g., negligible anomalies vs. critical defects).

\begin{figure}[t]
\centering
\includegraphics[width=1\linewidth]{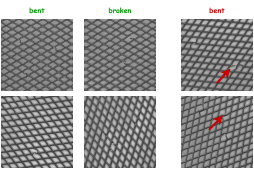}
\caption{Examples of misclassified samples in the MVTec-AD dataset. The first column displays bent samples, and the second column shows broken samples. However, the third column contains broken samples incorrectly labeled as bent, highlighting the need for dataset refinement.}
\label{fig:grid_mis}
\end{figure}

\subsection{Dataset Refinement for Anomaly Classification}

Existing anomaly detection datasets primarily focus on localization and often contain imprecise or mislabeled defect annotations. To establish a robust benchmark for anomaly classification, we refine two widely used datasets: MVTec-AD and VisA. To distinguish the modified dataset from the original one, we name them MVTec-AC and VisA-AC, where AC stands for Anomaly Classification.

\subsubsection{MVTec-AC}
\label{sec:mvtec_ac}
\begin{figure}[t]
\centering
\includegraphics[width=1\linewidth]{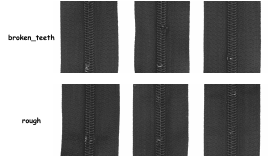}
\caption{Examples of \textit{broken\_teeth} and \textit{rough} anomaly classes in the MVTec-AD dataset. Despite their visual similarity, these anomalies are categorized into distinct classes, demonstrating the necessity for dataset refinement.}
\label{fig:zipper_merge}
\end{figure}

MVTec-AD~\cite{bergmann2019mvtec}, a widely used benchmark with 15 object categories, presents challenges for anomaly classification due to inconsistent labeling. To address these limitations, we introduce \emph{MVTec-AC}, which incorporates the following refinements:
\begin{itemize}
\item Correcting 36 misclassified samples across object categories (see Figure~\ref{fig:grid_mis}).
\item Merging four overlapping anomaly classes (see Figure~\ref{fig:zipper_merge}): poke and crack (capsule), cut and hole (carpet), thread\_side and thread\_top (screw), and broken\_teeth and rough (zipper)
\item Removing the \emph{toothbrush} category, which contains only one trivial anomaly class.
\item Excluding four 'combined' anomaly classes in the MVTec-AD dataset, as they group multiple anomalies and do not provide specific severity information, which is key for anomaly classification.
\end{itemize}
These refinements ensure a more structured and precise evaluation for anomaly classification.
\subsubsection{VisA-AC}
VisA~\cite{zou2022spot}, another widely used dataset, provides anomaly class information in an Excel file and comprises 12 object categories. However, directly restructuring the dataset leads to many anomaly classes with insufficient sample sizes. To ensure statistical relevance, we:
\begin{itemize}
\item Remove anomaly classes with fewer than 10 samples.
\item Merge four highly similar anomaly classes.
\item Correct three misclassified samples after manual review.
\end{itemize}
This results in \emph{VisA-AC}, a more suitable benchmark for evaluating industrial anomaly classification methods.

By integrating vision-based anomaly detection with multimodal LLM classification, VELM introduces a novel approach to anomaly characterization. Our method ensures computational efficiency, enhances interpretability, and supports adaptable classification criteria. Furthermore, through MVTec-AC and VisA-AC, we establish the well-structured benchmarks specifically designed for anomaly classification. These contributions lay the foundation for more effective and practical industrial anomaly analysis.

\section{Experiments}
\label{sec:exp}
\subsection{Evaluation Metrics}
To evaluate VELM, we utilize three key metrics: macro accuracy, macro F1-score, and Cohen's kappa. These metrics ensure a robust assessment of classification performance, accounting for class imbalances and agreement beyond chance.

\begin{table}[t]
\centering
\begin{tabular}{cccc}
\hline
            & \textbf{Acc}           & \textbf{F1}            & \textbf{Kappa}         \\ \hline
Echo        & 72.9          & -             & -             \\
MCAD        & 76.4          & -             & -             \\
VELM (ours) & \textbf{81.4} & \textbf{78.0} & \textbf{76.8} \\ \hline
\end{tabular}
\caption{Performance on the MVTec-AD dataset. Acc = Accuracy, Kappa = Cohen's Kappa. All metrics are reported as percentages}
\label{tab:VELMmvtec}
\end{table}

\begin{table*}[t]
\centering
\renewcommand{\arraystretch}{1.1}
\begin{tabular}{lccc|ccc|ccc|ccc}
\hline
              & \multicolumn{3}{c|}{\textbf{\begin{tabular}[c]{@{}c@{}}VELM\\ (Oracle+GPT-4o)\end{tabular}}} & \multicolumn{3}{c|}{\textbf{\begin{tabular}[c]{@{}c@{}}VELM\\ (DDAD+GPT-4o)\end{tabular}}} & \multicolumn{3}{c|}{\textbf{\begin{tabular}[c]{@{}c@{}}VELM\\ (PatchCore+GPT-4o)\end{tabular}}}          & \multicolumn{3}{c}{\textbf{\begin{tabular}[c]{@{}c@{}}VELM\\ (DDAD+GPT-4o-mini)\end{tabular}}}          \\ \cline{2-13} 
              & \textbf{Acc}                  & \textbf{F1}                 & \textbf{Kappa}                 & \textbf{Acc}                 & \textbf{F1}                  & \textbf{Kappa}               & \multicolumn{1}{c}{\textbf{Acc}} & \multicolumn{1}{c}{\textbf{F1}} & \multicolumn{1}{c|}{\textbf{Kappa}} & \multicolumn{1}{c}{\textbf{Acc}} & \multicolumn{1}{c}{\textbf{F1}} & \multicolumn{1}{c}{\textbf{Kappa}} \\ \hline
bottle        & 85.5                          & 85.2                        & 80.4                           & \textbf{86.7}                & \textbf{86.5}                & \textbf{86.7}                & 84.3                             & 82.9                            & 79.0                                & 74.7                             & 74.5                            & 66.3                               \\
cable         & 89.9                          & 83.4                        & 87.0                           & \textbf{86.3}                & 79.9                         & \textbf{86.3}                & \textbf{86.3}                    & \textbf{80.6}                   & 82.4                                & 72.7                             & 56.1                            & 64.1                               \\
capsule       & 86.6                          & 87.9                        & 83.2                           & \textbf{79.8}                & \textbf{80.2}                & \textbf{74.8}                & 69.7                             & 69.9                            & 62.0                                & 66.4                             & 66.4                            & 58.3                               \\
carpet        & 87.2                          & 86.8                        & 83.7                           & \textbf{82.9}                & \textbf{82.4}                & \textbf{82.9}                & 82.1                             & 82.3                            & 77.1                                & 81.2                             & 80.5                            & 76.1                               \\
grid          & 67.9                          & 58.7                        & 60.9                           & \textbf{66.7}                & 57.7                         & \textbf{66.7}                & 63.4                             & 55.6                            & 56.8                                & 65.4                             & \textbf{59.8}                   & 57.5                               \\
hazelnut      & 95.5                          & 94.6                        & 94.1                           & 91.8                         & 91.2                         & 89.2                         & \textbf{95.5}                    & \textbf{94.6}                   & \textbf{94.1}                       & 89.1                             & 87.8                            & 85.6                               \\
leather       & 88.7                          & 87.3                        & 86.3                           & 86.3                         & 84.8                         & 83.3                         & 73.4                             & 72.5                            & 66.9                                & \textbf{89.5}                    & \textbf{88.3}                   & \textbf{87.3}                      \\
metal nut     & 93.0                          & 92.9                        & 91.2                           & \textbf{90.4}                & \textbf{90.3}                & \textbf{90.4}                & 88.6                             & 88.4                            & 85.7                                & 77.2                             & 77.4                            & 71.4                               \\
pill          & 83.3                          & 79.8                        & 80.3                           & \textbf{79.2}                & \textbf{76.4}                & \textbf{75.3}                & 72.2                             & 70.4                            & 67.1                                & 59.7                             & 52.1                            & 52.0                               \\
screw         & 95.3                          & 80.5                        & 94.1                           & \textbf{83.9}                & \textbf{69.9}                & \textbf{79.7}                & 66.4                             & 56.5                            & 56.0                                & 69.1                             & 65.6                            & 61.2                               \\
tile          & 90.6                          & 86.6                        & 88.5                           & 89.7                         & 85.9                         & 87.4                         & \textbf{90.6}                    & \textbf{88.0}                   & \textbf{88.4}                       & 77.8                             & 69.0                            & 72.7                               \\
transistor    & 92.0                          & 83.8                        & 86.7                           & \textbf{89.0}                & \textbf{79.5}                & \textbf{81.4}                & 71.0                             & 64.9                            & 58.6                                & 78.0                             & 51.0                            & 62.7                               \\
wood          & 97.1                          & 96.8                        & 96.2                           & \textbf{88.2}                & 88.3                         & \textbf{84.8}                & \textbf{88.2}                    & \textbf{89.1}                   & 84.7                                & 85.3                             & 83.1                            & 80.9                               \\
zipper        & 77.0                          & 74.0                        & 72.1                           & \textbf{75.6}                & \textbf{73.5}                & \textbf{70.3}                & 59.3                             & 57.8                            & 50.5                                & 73.3                             & 69.8                            & 67.4                               \\ \hline
\textbf{Mean} & 87.8                          & 84.2                        & 84.6                           & \textbf{84.0}                & \textbf{80.3}                & \textbf{79.7}                & 78.1                             & 75.3                            & 72.1                                & 75.7                             & 70.1                            & 68.8                               \\ \hline
\end{tabular}
\caption{Comparative analysis of various VELM configurations on the MVTec-AC dataset. Each configuration combines different vision experts—Ground Truth, DDAD, and PatchCore—with multimodal large language models GPT-4o and GPT-4o-mini. Acc = Accuracy, Kappa = Cohen's Kappa. All metrics are reported as percentages}
\label{tab:VELMmacmvtec}
\end{table*}

\paragraph{Macro Accuracy} Since object categories contain varying numbers of anomaly classes, accuracy is first computed at the object level before aggregation. The macro accuracy is then obtained by averaging across all object categories:
\begin{equation}
    \text{Macro-Acc} = \frac{1}{|O|} \sum_{o \in O}\Bigg(\frac{1}{|\mathcal{C}_o|} \sum_{c \in \mathcal{C}_o} \frac{TP_c}{TP_c + FP_c + FN_c}\Bigg),
\end{equation}
where \( O \) is the set of all object categories, \( \mathcal{C}_o \) is the set of anomaly classes within object category \( o \), and \( TP_c \), \( FP_c \), and \( FN_c \) denote the number of true positives, false positives, and false negatives for class \( c \), respectively.

\paragraph{Macro F1-score} we utilize the macro F1-score to address class imbalances. For each anomaly class \( c \), we compute its F1-score. Next, the F1-score for an object category \( o \) is obtained by averaging over all its anomaly classes. Finally, the total macro F1-score is computed by averaging over all object categories:
\begin{equation}
    \text{Macro-F1} = \frac{1}{|O|} \sum_{o \in O}\Bigg(\frac{1}{|\mathcal{C}_o|} \sum_{c \in \mathcal{C}_o} F1_c\Bigg)
\end{equation}

\paragraph{Cohen’s Kappa} To evaluate classification performance while accounting for agreement by chance, we employ Cohen’s kappa coefficient, defined as:
\begin{equation}
    \kappa = \frac{p_o - p_e}{1 - p_e},
\end{equation}
where \( p_o \) is the observed agreement between the model predictions and ground truth, and \( p_e \) represents the expected agreement under random chance. This metric provides a more informative measure of classification reliability. 

These metrics provide a comprehensive evaluation of VELM across the MVTec-AC and VisA-AC datasets.

\subsection{Experimental Settings and Results}
In this section, we evaluate VELM by comparing it to the existing state-of-the-art in anomaly classification on the MVTec-AD dataset (Section~\ref{sec:exp_mvtec_ad}) and assessing its performance on the proposed MVTec-AC (Section~\ref{sec:exp_mvtec_ac}) and VisA-AC (Section~\ref{sec:exp_visa_ac}) datasets. For the multimodal LLM-based classification module, input images are resized to a resolution of 448×448 pixels, and the language model's temperature is set to zero. Each evaluation involves randomly selecting a reference normal image from the training set, along with a query image, an annotated image with localized anomalies provided by a vision expert, and a text prompt. These inputs are fed into the multimodal LLM, which outputs the anomaly class of the query image. We evaluate both the performance of the complete pipeline and the quality of the proposed MLLM-based classification module as a standalone component, obtaining the latter result through Oracle-based evaluations.

\subsubsection{Experiments on MVTec-AD}
\label{sec:exp_mvtec_ad}
We start by evaluating on the original MVTec-AD dataset. Despite the limitations of its annotations (see Section~\ref{sec:mvtec_ac}), evaluating on MVTec-AD allows us to compare VELM with two existing state-of-the-art anomaly classification methods: MCAD~\cite{li2024mcad}, a vision-based approach using relational knowledge distillation, and Echo~\cite{chen2025can}, which employs multiple language model components. VELM achieves an anomaly classification accuracy of $\textbf{81.4\%}$, surpassing Echo and MCAD by $9.5\%$ and $5\%$, respectively (Table~\ref{tab:VELMmvtec}).

\begin{table*}
\centering
\renewcommand{\arraystretch}{1.1}
\begin{tabular}{lccc|ccc|ccc|ccc}
\hline
              & \multicolumn{3}{c|}{\textbf{\begin{tabular}[c]{@{}c@{}}VELM\\ (Oracle+GPT-4o)\end{tabular}}} & \multicolumn{3}{c|}{\textbf{\begin{tabular}[c]{@{}c@{}}VELM\\ (DDAD+GPT-4o)\end{tabular}}} & \multicolumn{3}{c|}{\textbf{\begin{tabular}[c]{@{}c@{}}VELM\\ (PatchCore+GPT-4o)\end{tabular}}}          & \multicolumn{3}{c}{\textbf{\begin{tabular}[c]{@{}c@{}}VELM\\ (DDAD+GPT-4o-mini)\end{tabular}}}          \\ \cline{2-13} 
              & \textbf{Acc}                  & \textbf{F1}                 & \textbf{Kappa}                 & \textbf{Acc}                 & \textbf{F1}                  & \textbf{Kappa}               & \multicolumn{1}{c}{\textbf{Acc}} & \multicolumn{1}{c}{\textbf{F1}} & \multicolumn{1}{c|}{\textbf{Kappa}} & \multicolumn{1}{c}{\textbf{Acc}} & \multicolumn{1}{c}{\textbf{F1}} & \multicolumn{1}{c}{\textbf{Kappa}} \\ \hline
candle        & 89.9                          & 76.2                        & 83.6                           & \textbf{75.7}                & \textbf{66.1}                & \textbf{63.1}                & 71.6                             & 46.6                            & 44.5                                & 66.9                             & 42.4                            & 49.3                               \\
capsules      & 72.5                          & 64.7                        & 64.0                           & \textbf{67.5}                & \textbf{56.9}                & \textbf{55.8}                & 46.3                             & 26.1                            & 20.6                                & 56.9                             & 41.1                            & 43.0                               \\
cashew        & 73.1                          & 56.8                        & 65.6                           & 50.4                         & \textbf{29.8}                & \textbf{33.4}                & \textbf{52.9}                    & 27.6                            & 29.4                                & 46.2                             & 21.6                            & 28.0                               \\
chewinggum    & 91.2                          & 86.7                        & 87.8                           & 75.2                         & 63.1                         & 63.2                         & \textbf{77.9}                    & \textbf{67.8}                   & \textbf{67.3}                       & 66.4                             & 49.5                            & 50.0                               \\
fryum         & 92.1                          & 88.3                        & 88.8                           & 77.2                         & 71.9                         & 67.2                         & \textbf{82.5}                    & \textbf{78.6}                   & \textbf{74.5}                       & 66.7                             & 57.5                            & 52.0                               \\
macaroni1     & 81.1                          & 44.2                        & 72.3                           & \textbf{66.3}                & \textbf{34.2}                & \textbf{50.8}                & 58.4                             & 21.0                            & 23.0                                & 61.6                             & 25.2                            & 43.3                               \\
macaroni2     & 93.5                          & 88.0                        & 89.4                           & \textbf{65.3}                & \textbf{57.4}                & \textbf{46.9}                & 61.2                             & 35.2                            & 17.3                                & 60.0                             & 45.9                            & 38.8                               \\
pcb1          & 85.9                          & 77.0                        & 78.7                           & \textbf{75.4}                & \textbf{65.8}                & \textbf{64.5}                & 64.4                             & 47.7                            & 34.9                                & 68.1                             & 53.5                            & 54.0                               \\
pcb2          & 89.6                          & 80.5                        & 83.9                           & 65.4                         & 53.1                         & 41.1                         & \textbf{70.2}                    & \textbf{61.0}                   & \textbf{49.4}                       & 63.4                             & 44.8                            & 36.5                               \\
pcb3          & 85.1                          & 73.6                        & 77.6                           & \textbf{63.4}                & \textbf{55.3}                & \textbf{49.0}                & 60.3                             & 36.8                            & 24.8                                & 59.8                             & 49.7                            & 43.8                               \\
pcb4          & 98.6                          & 96.7                        & 96.9                           & 93.4                         & \textbf{87.4}                & 86.1                         & \textbf{94.4}                    & 90.6                            & \textbf{87.3}                       & 92.3                             & 84.1                            & 83.0                               \\
pipe\_fryum   & 99.3                          & 99.2                        & 99.1                           & 59.4                         & 60.0                         & 45.0                         & \textbf{81.9}                    & \textbf{83.9}                   & \textbf{76.0}                       & 53.6                             & 50.2                            & 37.5                               \\ \hline
\textbf{Mean} & 87.6                          & 77.7                        & 82.3                           & \textbf{69.6}                & \textbf{58.4}                & \textbf{55.6}                & 68.5                             & 51.9                            & 45.7                                & 63.5                             & 47.1                            & 46.6                               \\ \hline
\end{tabular}
\caption{Comparative analysis of various VELM configurations on the VisA-AC dataset. Each configuration combines different vision experts—Ground Truth, DDAD, and PatchCore—with multimodal large language models GPT-4o and GPT-4o-mini. Acc = Accuracy, Kappa = Cohen's Kappa. All metrics are reported as percentages}
\label{tab:VELMmacvisa}
\end{table*}

\subsubsection{Experiments on MVTec-AC}
\label{sec:exp_mvtec_ac}
To thoroughly evaluate VELM on MVTec-AC, we conduct experiments with different configurations:

\begin{itemize}
\item An assessment of the classification module alone, using ground truth anomaly segmentation boundaries as an ``oracle'', simulating a perfect vision anomaly detector.
\item A full pipeline evaluation with off-the-shelf vision anomaly detectors, testing two vision experts: DDAD~\cite{mousakhan2023anomaly} and PatchCore~\cite{roth2022towards}.
\item A comparison using a more lightweight LLM, GPT-4o-mini, alongside DDAD.
\end{itemize}
Table~\ref{tab:VELMmacmvtec} summarizes the results. Under ``oracle'' conditions, VELM achieves an anomaly classification accuracy of $87.8\%$. When using DDAD, performance drops to $84.0\%$ accuracy, $80.3\%$ F1-score, and $79.7\%$ Cohen’s kappa, while PatchCore with GPT-4o further lowers accuracy to $\textbf{78.1\%}$. Substituting GPT-4o with GPT-4o-mini leads to an additional decline, with DDAD + GPT-4o-mini achieving $\textbf{75.7\%}$ accuracy.

Despite the expected performance reduction in realistic scenarios, these results confirm that VELM remains effective across different anomaly detectors and LLMs, as long as the vision expert is sufficiently accurate.

\subsubsection{Experiments on VisA-AC}
\label{sec:exp_visa_ac}
Likewise, we conduct multiple evaluations on VisA-AC, testing different configurations. Table~\ref{tab:VELMmacvisa} summarizes the results. Under ``oracle'' conditions, VELM achieves \textbf{87.6\%} accuracy. When using DDAD as the vision expert, accuracy drops to \textbf{69.6\%}. PatchCore performs comparably, achieving \textbf{68.5\%} accuracy. Replacing GPT-4o with GPT-4o-mini leads to a further performance decrease, with DDAD + GPT-4o-mini obtaining \textbf{63.5\%} accuracy.

These results highlight the critical role of accurate anomaly localization. The performance drop when using DDAD and PatchCore suggests that noisy segmentation masks significantly impact classification accuracy.

\subsection{Anomaly vs. Defect}
In real-world applications, not all anomalies are defects—some deviations from the norm may be acceptable, while others require intervention. For example, in leather inspection, both water droplets and cuts may be considered anomalies, but only cuts represent critical defects that affect product quality. A practical anomaly detection system should differentiate between \textit{negligible anomalies} and \textit{defects} to support more informed decision-making in manufacturing and quality control.

To assess VELM's ability to make this distinction, we simulate a split using MVTec-AC. Specifically, we randomly designate $30\%$ of the anomaly class per object category as negligible anomalies and others as defects. This process is repeated five times with different random seeds to ensure robustness.
The results, presented in Table~\ref{tab:anodef}, show that VELM achieves a mean accuracy of 89.8\% in classifying each category (normal, anomaly, defect) versus the rest.  

These results highlight VELM’s potential for real-world anomaly classification, where prioritization of defects over minor deviations is essential. By enabling a finer-grained assessment of anomalies, VELM could support more precise quality control and automated decision-making in industrial applications.

\begin{table}[t]
\centering
\renewcommand{\arraystretch}{1.1}
\setlength{\tabcolsep}{6pt} 
\small 
\begin{tabular}{lllll}
\hline
           & \textbf{Normal} & \textbf{Anomaly} & \textbf{Defect} & \textbf{Total}         \\ \hline
bottle     & 1.0    & 87.3               & 85.9   & 91.1          \\
cable      & 1.0    & 77.1               & 86.6   & 87.9          \\
capsule    & 1.0    & 81.7               & 87.5   & 89.7          \\
carpet     & 1.0    & 74.7               & 84.3   & 86.3          \\
grid       & 1.0    & 57.3               & 77.5   & 78.3          \\
hazelnut   & 1.0    & 86.5               & 93.2   & 93.2          \\
leather    & 1.0    & 87.0               & 91.9   & 93.0          \\
metal\_nut & 1.0    & 96.4               & 92.8   & 96.4          \\
pill       & 1.0    & 82.1               & 85.5   & 89.2          \\
screw      & 90.2   & 90.1               & 88.5   & 89.6          \\
tile       & 1.0    & 83.4               & 95.3   & 92.9          \\
transistor & 1.0    & 78.0               & 88.0   & 88.7          \\
wood       & 1.0    & 84.5               & 91.1   & 91.9          \\
zipper     & 1.0    & 81.6               & 86.2   & 89.3          \\ \hline
Mean       & 99.3   & 82.0               & 88.2   & \textbf{89.8} \\ \hline
\end{tabular}
\caption{Accuracy (\%) of VELM in classifying normal samples, negligible anomalies, and critical defects on the MVTec-AC dataset.}
\label{tab:anodef}
\end{table}

\begin{table*}[t]
\centering
\renewcommand{\arraystretch}{1.1}
\resizebox{\textwidth}{!}{%
\begin{tabular}{lcccccc}
\toprule
\textbf{} & \textbf{VELM} & \textbf{VELM w/o RI} & \textbf{VELM w/o VP} & \textbf{VELM w/o ND} & \textbf{VELM w/o CS} & \textbf{VELM w/o AD} \\
\midrule
bottle     & \textbf{86.7} & 79.5          & 84.3          & 84.3          & 79.5          & 62.7          \\
cable      & 86.3          & 79.9          & \textbf{90.6} & 83.5          & 82.7          & 84.9          \\
capsule    & 79.8          & \textbf{82.4} & 75.6          & 80.7          & 79.8          & 65.6          \\
carpet     & 82.9          & 80.3          & \textbf{84.6} & 79.5          & 81.2          & 77.8          \\
grid       & 65.7          & \textbf{71.8} & 66.7          & 70.5          & 65.4          & 65.4          \\
hazelnut   & 91.8          & 91.8          & \textbf{92.7} & 90.9          & 91.8          & 92.8          \\
leather    & 86.3          & \textbf{89.5} & 88.7          & \textbf{89.5} & 86.3          & 87.1          \\
metal nut & 90.4          & 81.6          & 78.9          & \textbf{91.2} & 87.7          & 87.7          \\
pill       & \textbf{79.2} & 72.9          & 78.5          & 77.1          & 78.5          & 76.4          \\
screw      & \textbf{83.9} & 78.5          & 77.9          & 79.9          & 81.9          & 75.8          \\
tile       & 89.7          & 90.6          & \textbf{94.0} & 88.9          & 89.7          & 88.0          \\
transistor & 89.0          & 85.0          & 84.0          & \textbf{90.0} & \textbf{90.0} & 89.0          \\
wood       & 88.2          & 85.3          & \textbf{91.2} & 86.8          & 88.2          & 82.4          \\
zipper     & 75.6          & 73.3          & 69.6          & \textbf{76.3} & 74.1          & 64.4          \\
\midrule
\textbf{Mean} & \textbf{84.0} & 81.6          & 82.6          & 83.5          & 82.6          & 78.6          \\
\bottomrule
\end{tabular}
}
\caption{Ablation study on the different parts of the input prompts of VELM on the MVTec-AC dataset, in terms of anomaly classification accuracy. Abbreviations for the different parts of the prompt: RI = reference image, VP = visual prompt, ND = normal description, CS = classification strategy, AD = anomaly description.
The complete prompt has the best average accuracy, while the prompt without the description of the anomalies performs significantly worse. 
}
\label{tab:ablation}
\end{table*}

\subsection{Ablation Studies}
We conduct ablation studies to analyze the impact of input image configurations and prompt structure on classification performance.

We ablate the following elements of the VELM prompt: the normal reference image, red-line contour of the anomaly area (visual prompt), the textual description of the normal object, a description of the classification strategy, and descriptions of the anomalies.  

Table~\ref{tab:ablation} shows that using a reference image and visual prompts improves classification accuracy. Removing these elements reduces performance to $81.6\%$ and $82.6\%$, respectively. Text-based inputs are equally crucial; omitting anomaly descriptions significantly lowers accuracy. These findings demonstrate that a combination of both visual and textual inputs is essential for maximizing classification accuracy. The complete prompt structure yields the best performance, reinforcing the importance of multimodal inputs in effectively guiding VELM’s anomaly classification.

\section{Conclusion}
\label{sec:conc}
We introduced the first anomaly classification framework designed explicitly for anomaly classification using multimodal large language models (MLLMs). Unlike traditional approaches, our method requires no task-specific training while achieving high classification accuracy, up to \textbf{84\%}. We analyze the key components contributing to its success through ablation studies, demonstrating the importance of both visual and textual inputs. Additionally, we explore real-world applications by simulating scenarios where distinguishing defects from minor anomalies is crucial for automated decision-making. By bridging vision models with LLMs, we demonstrate a practical and flexible approach to anomaly classification.

To support evaluation, we refine the class annotations of MVTec-AD and VisA, addressing inconsistencies in existing anomaly detection datasets. These benchmarks establish a reliable foundation for assessing classification performance in real-world inspection scenarios.

\subsection{Future Work and Limitations}

A key limitation is its reliance on a closed set of user-defined classes. Extending the framework to handle open-set anomalies would improve adaptability. Additionally, integrating feedback between the vision expert and classification module could enhance robustness. Exploring these directions will further refine automated anomaly classification for industrial inspection.

{
    \small
    \bibliographystyle{ieeenat_fullname}
    \bibliography{main}
}


\end{document}